# Machine learning modeling of hit-into-play probability and quantification of pitch sequence effects

Ryota Takamido[1] and Hiroki Nakamoto[2]

1 Sports Innovation Organization, National Institute of Fitness and Sports in Kanoya, Kanoya, Kagoshima 891-2393, Japan

2 Faculty of Physical Education, National Institute of Fitness and Sports in Kanoya, Kanoya, Kagoshima 891-2393, Japan

Corresponding Author: Ryota Takamido

Street Address, Kanoya, Kagoshima, 891-2393, Japan

Email address: rtakamido@nifs-k.ac.jp

## Author contributions

RT and HN conceived and designed the study. RT collected and prepared the data, and performed the analysis. RT and HN contributed to the interpretation of the results. RT drafted the manuscript. HN critically revised the manuscript. Both authors read and approved the final manuscript.

## Statement and Declarations

### Declaration of conflicting interest

The author(s) declared no potential conflicts of interest with respect to the research, authorship, and/or publication of this article

## Funding statement

This study was supported by the Japan Society for the Promotion of Science under Grant JP25K21018. The funders played no role in the study design, data collection and analysis, decision to publish, or manuscript preparation.

## Ethical approval and informed consent statements

Ethical approval was not required for this study because it used only publicly available datasets and did not involve intervention with, interaction with, or collection of private or identifiable information from human participants.

## Data availability statement:

The source code for the analysis and datasets used in this study are publicly available on GitHub (https://github.com/takamido/Pitch_sequence_analysis).

## Consent for publication

Not applicable. This study used publicly available datasets and did not involve recruitment of or interaction with human participants.

## Abstract

Although pitch sequencing is a central topic in baseball analytics, previous studies have primarily focused on improving predictive performance for the final pitch or its outcome, leaving the role of preceding pitches insufficiently examined. To address this issue, this study conducted a machine learning–based analysis to quantify the importance of preceding pitches and deepen the tactical understanding of baseball. A Transformer-based model was designed and trained to predict whether a target pitch would result in an in-play outcome or a swinging strikeout. Based on the model, two analyses were conducted: an ablation study to evaluate the contribution of preceding pitches to predictive performance and a replacement analysis to quantify changes in predicted in-play probability when individual pitches were replaced by alternatives. The results suggest that hitting outcome predictions mostly depended on the attributes of the final pitch, and including preceding pitches had little effect on the model's predictive performance. Moreover, the effect of pitch replacement on the model's probability output for preceding pitches was approximately 0.11–0.14, compared with 0.44–0.45 for the final pitch. Therefore, preceding pitches played only a supplementary role in slightly modifying the predicted hit-into-play probability, although even such small effects may be practically meaningful.

**Keywords**: Pitch sequence, Baseball, Machine learning, Transformer

## 1. Introduction

The pitch sequence, which refers to the strategic order of pitch types and locations selected by the pitcher and catcher, is considered an important factor in baseball games. Because human perception is highly dependent on the sequence of stimuli (Cicchini et al., 2018; Cicchini et al., 2024), even when a pitcher throws the same pitch, the batter's perception of that pitch may vary depending on the velocity, location, spin axis, and other physical characteristics of the preceding pitch. Moreover, even if a pitcher throws the best pitch, the risk of being hit increases when the pitch is easily anticipated by the batter based on prior pitch history (Gray and Cañal-Bruland, 2018). Therefore, although hitting outcomes depend heavily on characteristics of the final pitch, such as its location and velocity, the effectiveness of that pitch may not be determined solely by its own characteristics but also by those of preceding pitches.

From a statistical perspective, predicting hitting outcomes from a given pitch sequence requires accounting for nonlinear and sequence-dependent relationships involving not only pitch-specific information, such as velocity and location, but also contextual factors, such as the count and game situation. Therefore, as summarized in Table 1, recent analytics studies have employed deep learning models, high-dimensional Bayesian models, and other advanced statistical approaches to capture these complex relationships and predict hitting outcomes or optimize pitch-sequence strategies. These studies have led to many important practical findings, as shown in the "Main practical insights" column of Table 1.

Despite these advances, a key limitation of previous studies is the insufficient examination of the importance of preceding pitches before the final pitch. Although many previous studies have adopted statistical models that account for a series of pitches, these studies have primarily focused on improving predictive performance for the final pitch or its outcome (Kneita, 2025; Lee, 2022; Bock, 2015; Hamilton et al., 2014), rather than understanding the role of preceding pitches in performing the prediction task. Therefore, many questions regarding the preceding pitches remain unclear, including the following examples: (1) The extent to which preceding pitches contribute to predicting the final-pitch outcome. (2) How far back in the pitch sequence preceding pitches provide meaningful information. (3) Which attributes of preceding pitches are important for predicting outcomes. (4) Whether the importance of preceding pitches varies across seasons. Addressing these questions is expected to deepen the tactical understanding of baseball and provide new practical insights.

Based on this background, this study conducted analyses focusing on preceding pitches.

Specifically, to answer the aforementioned questions, we performed two analyses using a predictive machine learning (ML) model that takes pitch sequence and game contextual information as input and outputs the in-play (or swinging strikeout) probability. The first was an ablation study examining how the model's predictive performance changed when specific information was excluded. This analysis directly quantifies the contribution of information from each pitch in the sequence to the prediction of the hitting outcome. The second was a replacement analysis examining how the model output changed when a specific pitch within the sequence was replaced with an alternative pitch. This analysis quantifies the relative importance of each pitch within the sequence to the model output. Although these analyses do not directly test practically meaningful hypotheses, we position this study as an exploratory, data-driven investigation to provide an initial basis for understanding the role and importance of preceding pitches.

Table 1. Summary of related studies on pitch sequence analysis. Representative recent studies published from 2015 to 2026 are shown.

| Study | Model | Task | Main practical insights |
|---|---|---|---|
| Kneita (2025) | Transformer-based sequence modeling | Pitch outcome prediction and optimization | Importance of sequential pitch context. Importance of pitcher-strength alignment |
| Melville et al. (2023) | RNN-based predictive model and Game-theoretic analysis | Pitch-sequence optimization | Strategic value of pitch tunneling. Importance of command-aware targeting |
| Nakahara et al. (2023) | Causal analysis using propensity scores | Causal effect estimation of fastball location | Effectiveness of the outside pitches |
| Lee (2022) | Ensemble deep neural networks | Pitch type and location prediction | Greater predictability in pitcher-disadvantaged situations |
| Prasad (2021) | Directed graph embeddings and Gaussian mixture clustering | Pitch-sequence clustering and pitcher-style analysis | Importance of pitch ordering beyond pitch selection. Identification of setup and knockout pitches |
| Healey & Zhao (2020) | Kernel regression. with Earth Mover's Distance | Pitch-frequency optimization | Importance of pitch distribution for improving strikeout rate |
| Healey (2019) | Bayesian nonparametric estimation | Pitch value estimation | Importance of pitch location in determining pitch value |
| Sidle & Tran (2018) | Random forest classifier | Pitch-type prediction | High predictability in batter-favored counts. Recent pitch history informs next-pitch prediction |
| Healey & Zhao (2017) | Multiple linear regression | Strikeout-rate prediction | Importance of pitch-to-pitch changes in velocity and movement |
| Swartz et al. (2017) | Random forest regression | Pitch-quality evaluation | Effectiveness of low-and-away pitches. Importance of count for pitch quality |
| Bock (2015) | Support vector machine classifier | Pitch-type prediction | Pitch-type predictability linked to long-term performance |

## 2. Method

### 2.1 Problem definition and ML model design

The problem considered in this study is the prediction of swing outcomes under the following condition: after reaching two strikes, the batter attempted to swing and the plate appearance ended as a result. Except for some specific cases such as catcher interference, the final outcome under this condition is limited to either a swinging strikeout or a ball put into play in fair territory. Here, the task of the ML model in this study was to predict the swing outcome, with a ball hit-into-play coded as 1.0 and a swinging strikeout as 0.0. If the ML model is well calibrated, its probability output corresponds to the expected in-play probability on a scale from 0.0 to 1.0. From a practical perspective, reducing the in-play rate (increasing the swinging-strike rate) is associated with lower probabilities of allowing baserunners and runs (Bradbury, 2007).

The reason for restricting the analysis to two-strike situations was to keep the condition that batters would generally attempt to swing at pitches within the strike zone. Moreover, because the value of a foul ball, such as whether it represents a "missed opportunity on a hittable pitch" or a “successful escape from a difficult pitch,” is difficult to quantify, we first adopted the simplified condition described above to examine the importance of preceding pitches.

Figure 1 illustrates the conceptual diagram of the ML model architecture used in this study. The ML model in this study was designed to receive pitch sequence and game context information as input and output the probability that the final pitch in the sequence resulted in a strikeout (0.0) or a ball hit-into-play in fair territory (1.0). To distinguish the contribution of game state and batter characteristics to hitting outcome prediction from that of pitch-sequence information, contextual information was also included as model input.

Transformer architecture (Vaswani et al., 2017) was used as the base predictive model. This selection was based on previous findings that the transformer achieved the best performance in pitch outcome prediction tasks (Kneita, 2025). The Transformer is a representative ML model used for sequential data processing. One of the advantages of this model is its strong capacity to model long-range dependencies in time-series data because of its attention-based architecture (Zhao et al., 2026). Because this study focuses on the application of the Transformer model, detailed descriptions of the model architecture are omitted; readers can refer to the original paper (Vaswani et al., 2017).

During the input data processing, the two types of input information were processed through separate pathways (Figure 1). The pitch sequence information was processed using a transformer. Owing to the aforementioned characteristics of the Transformer, this pathway is expected to effectively capture the relationships among pitches within a sequence, including complex variations in pitch location, velocity, and pitch type. The pitch sequence pathway comprises a linear projection layer, learnable positional embeddings, and Transformer Encoder blocks. Each Transformer Encoder block consists of multi-head self-attention, a feed-forward network, residual connections, and layer normalization. The resulting sequence representation is aggregated using masked mean pooling and passed through a dense layer. Concurrently, context information was processed using a separate dense layer. The pitch sequence and context representations are then concatenated and passed through a final dense layer, followed by a sigmoid output unit that produces the predicted in-play probability. Finally, because the analyses in this study (Analysis 2) were based on the model’s probability outputs, Platt scaling was applied to improve their calibration (Platt, 1999). Specifically, a logistic regression model was fitted to the logit-transformed raw predicted probabilities and the corresponding binary outcomes. The fitted calibration function was then used to transform the raw model outputs into calibrated predicted probabilities.

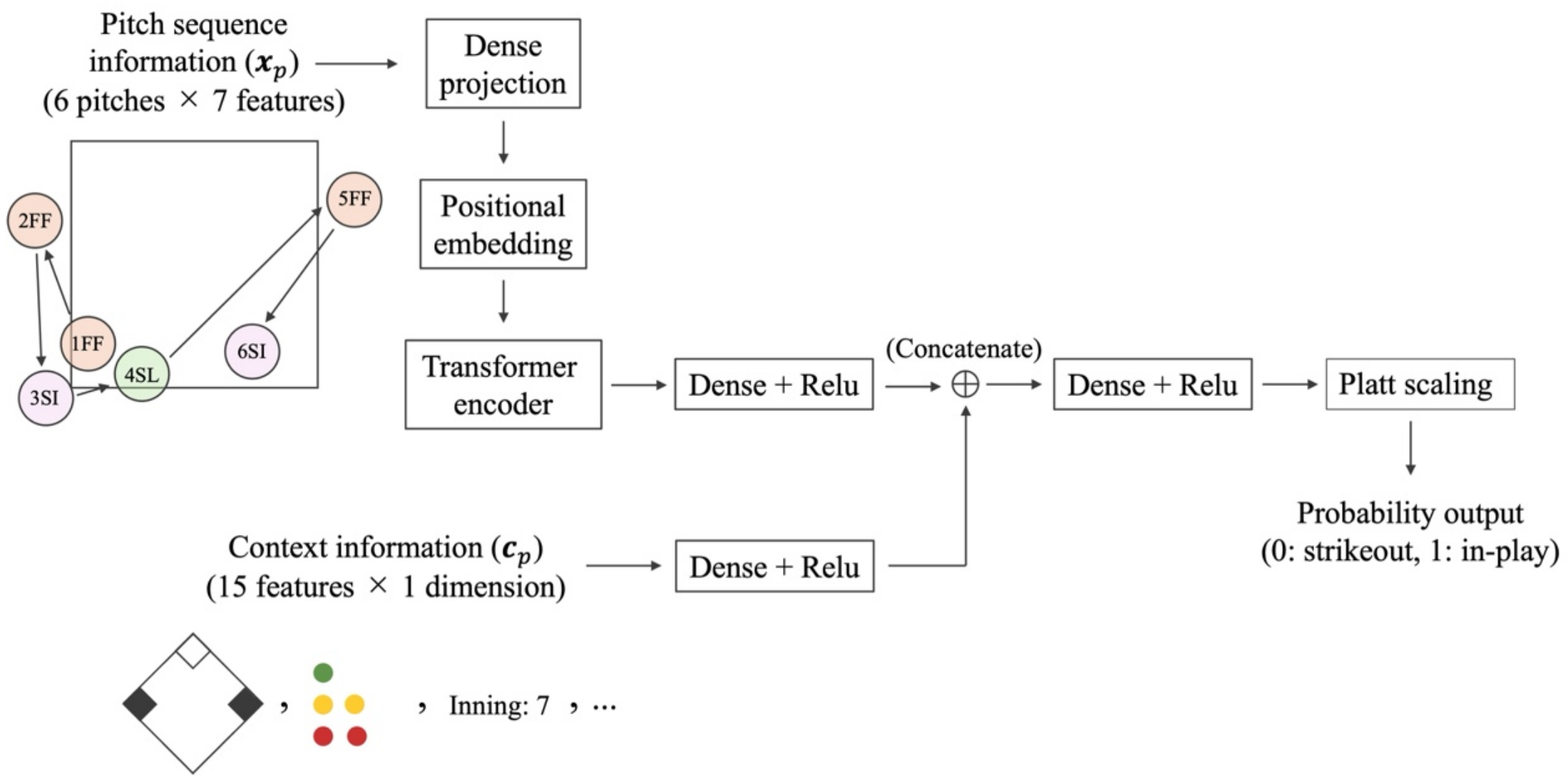

Figure 1. Conceptual diagram of the ML model architecture used in this study.

## 2.2 Dataset

Pitch-level Statcast data from the 2018 to 2025 regular season, excluding the shortened 2020 season, were downloaded using the Python package *pybaseball*. The pitches included in the analysis were screened based on the following criteria: (1) the pitches were thrown by pitchers who met the minimum innings pitched requirement in each target season, (2) the plate appearance ended after two strikes with either a swing out or a ball put into play in fair territory.

One analysis sample in the ML model was defined at the plate appearance level. For each plate appearance, the model receives the input features and outputs the in-play probability. Specifically, the input features consisted of two types of information: a pitch sequence and contextual information. The pitch sequence comprised the last six pitches of each plate. Seven features were used for each pitch: the horizontal plate location, vertical plate location, effective speed, release spin rate, spin axis, horizontal movement, and vertical movement. To standardize the sequence length, only the last six pitches were retained for plate appearances containing more than six pitches, whereas zero padding was applied at the beginning of the sequence for those containing fewer than six pitches. Consequently, each input pitch sequence is represented as a 6 × 7 matrix. The vertical plate location was normalized using batter-specific strike zone boundaries, and the horizontal movement and location were reversed for left-handed batters to align the coordinate system.

Contextual information included ball count, number of outs, inning, top/bottom inning indicator, base-runner indicators, score differential from the batting team's perspective, and the number of times the pitcher faced the batting order. These variables reflect the game's state at the final pitch. Furthermore, to account for the characteristics of the opposing batter, the batter's previous season statistics were included as contextual inputs: strikeout rate, home-run rate, isolated power, batting average, and OPS. For opposing batters without previous season statistics, the values were 10% worse than the average of available batters. Consequently, each context's input information was represented as a $15 \times 1$ matrix.

As a result of these procedures, a dataset of 100,669 samples involving pitches thrown by 265 pitchers was developed. More detailed summaries of the developed dataset, such as breakdowns by pitch type and the sample-selection process, are provided in the Supplementary Information (Tables S1–S5).

### 2.3 Base model training and testing

We first trained and tested a base model using all pitch-sequence and contextual information, and used its performance as the reference for the subsequent analyses. For model training and testing, 16,487 samples corresponding to the 2025 season were used as the test dataset, and the remaining 84,182 samples were used as the training dataset. Within the training dataset, 10% of the samples were used for validation.

The Transformer architecture used in this study consisted of an embedding dimension of 256, four attention heads, a feed-forward dimension of 512, two Transformer encoder blocks, and 64 units in the dense layer after concatenation. The dense-layer size of the contextual information processing pathway was set to 32. The model parameters were determined using hyperparameter tuning. Because a detailed tuning procedure is beyond the scope of this study, it is provided in the Supplementary Information.

The model was trained for up to 200 epochs with early stopping based on the validation AUC with a patience of 10 epochs. The learning rate and batch size were set at $1\times10^{-4}$ and 512, respectively. Binary cross-entropy was used as the loss function, and the Adam optimizer was used for the optimization. The classification threshold was set at 0.50. After model training, Platt scaling was applied to calibrate the predicted probabilities. A logistic regression model was fitted using the validation dataset, with the logit-transformed raw predicted probabilities as the input and the corresponding binary outcomes as the target. The fitted calibration model was then applied to the raw predicted probabilities of the test dataset and obtained the calibrated probabilities.

The model performance was evaluated on the test set using accuracy, precision, recall, and F1 score. To assess the validity of the probability estimates, the Brier score and Expected Calibration Error (ECE) were also calculated, and calibration curves were generated. ECE was calculated using 10 equal-width probability bins between 0.0 and 1.0. The uncertainty of these evaluation metrics was assessed using 1,000 bootstrap resamples. In each bootstrap iteration, observations in the test set were resampled with replacement at the plate-appearance level, and 95% confidence intervals were calculated. To evaluate the effect of introducing Platt scaling, we also assessed how the Brier score and ECE changed with and without Platt scaling.

Furthermore, we also examined how various factors affected the model's predictive performance and calibration. First, to evaluate the model's generalizability across different seasons, we performed an evaluation using an expanding-window approach. Specifically, four training and test settings were established: 2018–2019 for predicting 2022, 2018–2022 for predicting 2023, 2018–2023 for predicting 2024, and 2018–2024 for predicting 2025, with the shortened 2020 season excluded from all training periods. The 2021 season was also excluded from testing because the contextual information included batter statistics from the previous season. In each setting, the model was retrained while keeping the aforementioned parameters fixed, and the same evaluation metrics described above were applied.

Additionally, using the model trained in the setting where 2025 served as the test set, we examined variations in model calibration across ball counts, final-pitch types, final-pitch locations inside or outside the strike zone, batter handedness, and pitcher strikeout rate (K%). Because the subsequent analyses were based on the model's probability outputs, this evaluation was intended to comprehensively examine factors that could affect the reliability of these outputs. The model was evaluated separately across the categories within each factor. For example, for strike-zone location, the test set was divided into pitches inside and outside the strike zone, and the calibration metrics were compared between the two groups. For the pitch-type analysis, only pitch types with at least 200 samples were included to ensure an adequate sample size. For pitcher K%, the target pitchers were divided into tertiles based on their 2025 season statistics, forming Lower third, Middle third, and Upper third groups. Calibration metrics were then compared across the three K% groups to examine whether the model systematically overestimated or underestimated contact probabilities for pitchers with relatively high or low strikeout ability.

### 2.4 Analysis 1: Ablation study

To evaluate the importance of pitch-sequence information, we conducted an ablation study comparing the models listed in Table 2. Specifically, we first defined conditions in which information from each individual pitch position (only the 1st, 2nd, 3rd, 4th, 5th or 6th pitch counted back from the final pitch; Last1 is the final pitch) was used together with contextual information and trained the model under each condition. We also evaluated models using progressively longer sequences consisting of the last 1–2, 1–3, 1–4, or 1–5 pitches together with contextual information. Comparisons between these ablation models and the Full model allow us to directly evaluate the contribution of each pitch in the sequence to predictive performance. If the model relies only on the final pitch to predict the hitting outcome, removing the final pitch information would result in a substantial decrease in predictive performance, whereas adding information from preceding pitches would provide little benefit.

Furthermore, we defined conditions using only the pitches preceding the final pitch, using the final pitch together with one randomly selected preceding pitch, shuffling the order of the entire pitch sequence, shuffling the order of the preceding pitches while keeping the final pitch fixed, or using only contextual information, and trained and evaluated the model under each condition in the same manner. If the order of pitches is not important and only the fact that a pitch was thrown previously matters, shuffling the pitch order would not be expected to affect predictive performance.

Finally, as simpler baseline models, we included a logistic regression model and a constant baseline that assigned the in-play rate observed in the training data as the same probability to all test samples. For the logistic regression model, the input consisted of a one-dimensional vector of pitch features from up to six pitches concatenated with the contextual features. All features were standardized based on the training data, and the model was trained using the *LogisticRegression* implementation in scikit-learn. The maximum number of iterations was set to 1,000. The trained model was then used to estimate the in-play probability for each test sample.

To assess the stability of the results across seasons, these analyses were conducted using

four test seasons described above (2022, 2023, 2024 and 2025).

Table 2. List of models included in the ablation study.

| Model name | Brief explanation |
| --- | --- |
| Full | Base model with access to all input information. Evaluation results from Section 2.3 were used for this model. |
| Last1, Last2, Last3, Last4, Last5, Last6 | Model with access to only the 1st, 2nd, 3rd, 4th, 5th, or 6th pitch counted back from the final pitch (Last1 = final pitch) and context information. |
| Last1–2, Last1–3, Last1–4, Last1–5 | Model with access to only the last 1–2, 1–3, 1–4, or 1–5 pitches and context information. |
| Without final pitch | Model with access to all preceding pitches (excluding the final pitch) and contextual information. |
| Final pitch plus one | Model using the final pitch and one randomly selected preceding pitch. |
| All-shuffle | Model with all pitches randomly shuffled in order. |
| Partial-shuffle | Model with the final pitch fixed and the order of all preceding pitches randomly shuffled. |
| Context only | Model using only contextual information. |
| Logistic regression | Logistic regression model using all pitch-sequence and contextual information. |
| Constant | Model that predicts the same in-play probability for all samples based on the in-play rate in the training data. |

**2.5 Analysis 2: Replacement analysis**

Because Analysis 1 focused on binary classification performance, it may not fully capture the effects of preceding pitches if those effects were too small to change the output label. Therefore, in Analysis 2, the changes in output probability were directly quantified when a specific pitch within the sequence was replaced with an alternative pitch. This analysis was conducted using the datasets for which calibration was stable based on the results of the base model evaluation (those with 2023, 2024, and 2025 as the test years). The trained full model using all input information with Platt scaling applied was used.

Specifically, for each test sample, the original plate-appearance context and pitch sequence were retained, whereas only a specific pitch within the sequence was replaced with an alternative. To improve the practical relevance of the simulation conditions, only pitch sequences in which the original final pitch was located within the strike zone were included, and among the preceding pitches, only pitches thrown outside the strike zone and taken by the batter were selected for replacement. This assumes that batters with two strikes respond consistently to replacement pitches, swinging at final pitches in the strike zone and taking preceding pitches outside the strike zone. Because priority was given to maintaining realistic assumptions, the replacement conditions for the final and preceding pitches were not completely identical. This should therefore be considered when interpreting the results.

For the replacement procedure, two conditions were applied: an artificial-pitch condition and a representative observed-pitch condition. While the former broadly examines a range of possible replacements, the latter restricts replacements to pitches that were actually thrown by the pitcher.

In the artificial-pitch condition, the mean pitch velocity, spin rate, spin axis, and movement for each pitch type were calculated for each pitcher, and the corresponding features of the target pitch were replaced with these values. For pitch location, the strike zone and ball zone were each divided into cells as shown in Figure 2, and the center of each cell was used as a simulated pitch location. For all combinations of pitch type and pitch location, the target pitch was replaced with the corresponding candidate pitch while the remaining pitch-sequence information was retained. The modified sequence was then re-entered into the model, and the resulting output was evaluated. Relative to the original model output probability, the maximum increase and maximum decrease were calculated for each target pitch. For each pitcher, the min–max range of output changes was averaged for each pitch-sequence position, and these pitcher-level averages were then averaged across pitchers. Changes resulting from replacing only pitch type or only pitch location were also summarized to identify which attributes showed greater sensitivity in the trained model.

In the observed-pitch condition, a representative pitch was selected from the observed pitches for each pitcher, pitch type, and pitch-location region. Representative pitches were selected using a rank-center method. First, pitches within each grid cell in Figure 2 were ranked according to seven features: pitch velocity, spin rate, spin axis, horizontal and vertical movement, horizontal pitch location, and normalized vertical pitch location. For each pitch, the absolute difference between its rank for each feature and the median rank was calculated and summed across the seven features. The observed pitch with the lowest resulting rank-center score was selected as the representative pitch for that pitcher, pitch type, and region. Representative pitches were selected only from grid cells containing at least five observed pitches and cells that did not meet this criterion were excluded from replacement. The results were summarized using the same procedure as in the artificial-pitch condition.

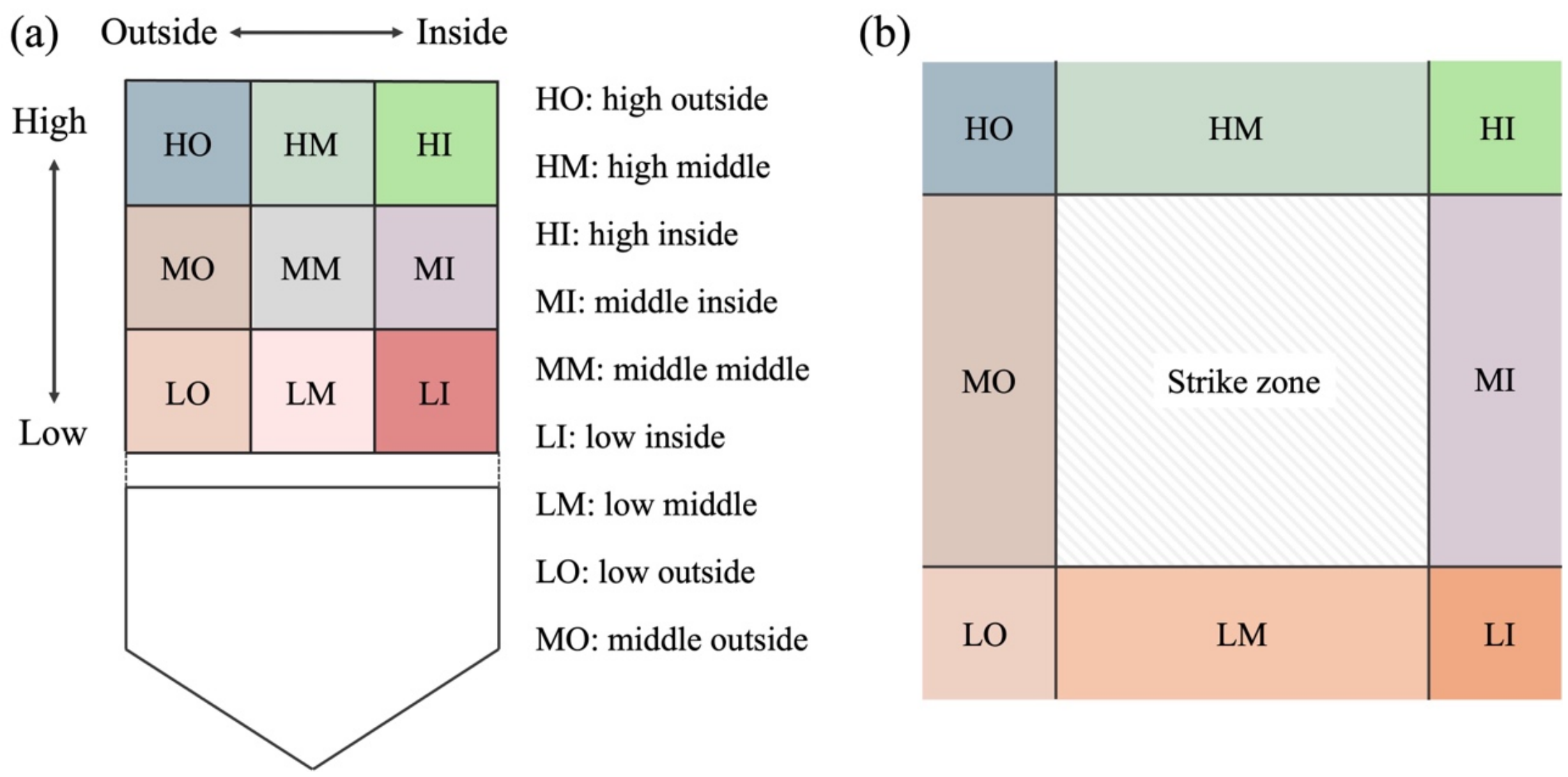

Figure 2. Candidate replacement locations: (a) the strike zone divided into a 3 × 3 grid and (b) ball zone. Although the ball zone is shown within a limited plotting range in this figure, it is not bounded in either the horizontal or vertical direction in the analysis.

## 3. Results

### 3.1 Base model training and testing

As a result of testing the base model, the model achieved an accuracy of 0.750 (95% CI: 0.743–0.756), precision of 0.761 (0.752–0.768), recall of 0.850 (0.843–0.857), F1 score of 0.803 (0.796–0.808), and ROC AUC of 0.807 (0.800–0.813). While Platt scaling did not improve the Brier score, which remained 0.170 before and after calibration, the ECE improved from 0.024 to 0.018. Table 3 presents the confusion matrices of the base model. Although some prediction errors remained, the trained model generally captured the tendency of pitch sequences to result in either in-play outcomes or swing-outs.

Table 4 shows the comparison of model predictive performance and calibration across different combinations of training and test years. The model showed relatively stable results when data from 2023–2025 were used as the test datasets. In contrast, both metrics tended to be slightly poorer when the 2022 data were used as the test dataset; this may reflect the relatively limited sample size available for training in relation to the complexity of the model.

Table 5 shows the calibration results for each factor considered in this study, and Figure 3 presents the calibration curves generated based on these results. Although the model showed stable predictive performance for most factors, with Brier scores below 0.20 and ECE values below 0.05, calibration tended to deteriorate for several factors. Specifically, when the 2022 data were used as the test dataset, the model tended to underestimate the in-play probability (Figure 3 (b)). Therefore, in the subsequent Analysis 2, only the datasets using 2023–2025 as the test years were included. Additionally, in the comparison across pitch types, the Fastball showed the poorest calibration (Figure 3 (d)). This may be because fastballs are thrown by all pitchers and therefore show greater individual variability, making calibration more difficult. Moreover, the underestimation at low predicted probabilities within the strike zone (Figure 3 (e)) may be due to the relatively small number of samples in this range. Finally, regarding pitcher ability, the model tended to underestimate the in-play probability for pitchers with lower K% (Figure 3 (f)). This may be because factors other than pitch attributes, such as pitch predictability to opposing batters (Takamido et al., 2026), also influence hitting outcomes in practical settings. Although the model showed stable calibration on average (Figure 3 (a)), these factors affecting calibration should be considered when interpreting the subsequent results.

Table 3. Confusion matrix of the base model using the classification threshold of 0.50, with the 2025 data used as the test dataset.

| | Predicted in-play | Predicted swing-out |
|---|---|---|
| Actual in-play | 8404 | 1482 |
| Actual swing-out | 2644 | 3957 |

Table 4. Comparison of model performance across different combinations of training and test years. Values in parentheses indicate 95% confidence intervals calculated using bootstrap sampling.

| Test year (Train year) | Accuracy | Precision | Recall | F1-score | ROC AUC | Brier score | ECE |
|---|---|---|---|---|---|---|---|
| 2022 (2018–2019) | 0.732 (0.724–0.738) | 0.776 (0.767–0.785) | 0.786 (0.777–0.794) | 0.781 (0.774–0.788) | 0.791 (0.784–0.799) | 0.180 (0.177–0.183) | 0.060 (0.053–0.066) |
| 2023 (2018–2022) | 0.750 (0.743–0.757) | 0.761 (0.752–0.770) | 0.837 (0.829–0.845) | 0.797 (0.791–0.803) | 0.813 (0.805–0.820) | 0.170 (0.167–0.173) | 0.016 (0.011–0.023) |
| 2024 (2018–2023) | 0.743 (0.737–0.750) | 0.757 (0.750–0.765) | 0.846 (0.840–0.853) | 0.799 (0.793–0.805) | 0.797 (0.791–0.804) | 0.174 (0.171–0.177) | 0.020 (0.014–0.026) |
| 2025 (2018–2024) | 0.750 (0.743–0.756) | 0.761 (0.753–0.769) | 0.850 (0.843–0.857) | 0.803 (0.797–0.808) | 0.807 (0.800–0.814) | 0.170 (0.167–0.173) | 0.018 (0.013–0.025) |

Table 5. Calibration results for each factor. Results are shown for the analysis in which the 2025 data were used as the test dataset.

| Factor | Group | N | Brier score (95% CI) | ECE (95% CI) |
|---|---|---|---|---|
| Count | 0-2 | 3193 | 0.177 (0.170–0.184) | 0.017 (0.015–0.034) |
| | 1-2 | 5170 | 0.173 (0.168–0.179) | 0.024 (0.017–0.036) |
| | 2-2 | 4917 | 0.166 (0.161–0.172) | 0.023 (0.017–0.035) |
| | 3-2 | 3207 | 0.165 (0.158–0.172) | 0.021 (0.017–0.039) |
| Pitch type | Fastball | 4711 | 0.184 (0.178–0.189) | 0.070 (0.059–0.083) |
| | Changeup | 2695 | 0.194 (0.186–0.201) | 0.037 (0.029–0.057) |
| | Slider | 2040 | 0.162 (0.153–0.171) | 0.030 (0.022–0.051) |
| | Sinker | 1839 | 0.130 (0.119–0.140) | 0.016 (0.016–0.038) |
| | Sweeper | 1684 | 0.161 (0.152–0.170) | 0.023 (0.020–0.048) |
| | Curveball | 1240 | 0.154 (0.143–0.165) | 0.041 (0.030–0.069) |
| | Split | 775 | 0.193 (0.180–0.207) | 0.044 (0.034–0.083) |
| | Cutter | 749 | 0.179 (0.165–0.194) | 0.027 (0.026–0.067) |
| | Knuckle Curve | 648 | 0.136 (0.121–0.149) | 0.025 (0.026–0.064) |
| Strike zone | Inside | 8981 | 0.162 (0.158–0.166) | 0.037 (0.029–0.045) |
| | Outside | 7506 | 0.180 (0.176–0.184) | 0.013 (0.010–0.024) |
| Batter side | Left-handed | 6894 | 0.166 (0.161–0.170) | 0.018 (0.013–0.029) |
| | Right-handed | 9593 | 0.173 (0.169–0.177) | 0.021 (0.015–0.030) |
| Pitcher K% | Lower third | 5177 | 0.162 (0.157–0.167) | 0.049 (0.039–0.059) |
| | Middle third | 5433 | 0.172 (0.167–0.176) | 0.022 (0.017–0.035) |
| | Upper third | 5877 | 0.177 (0.171–0.182) | 0.015 (0.012–0.030) |

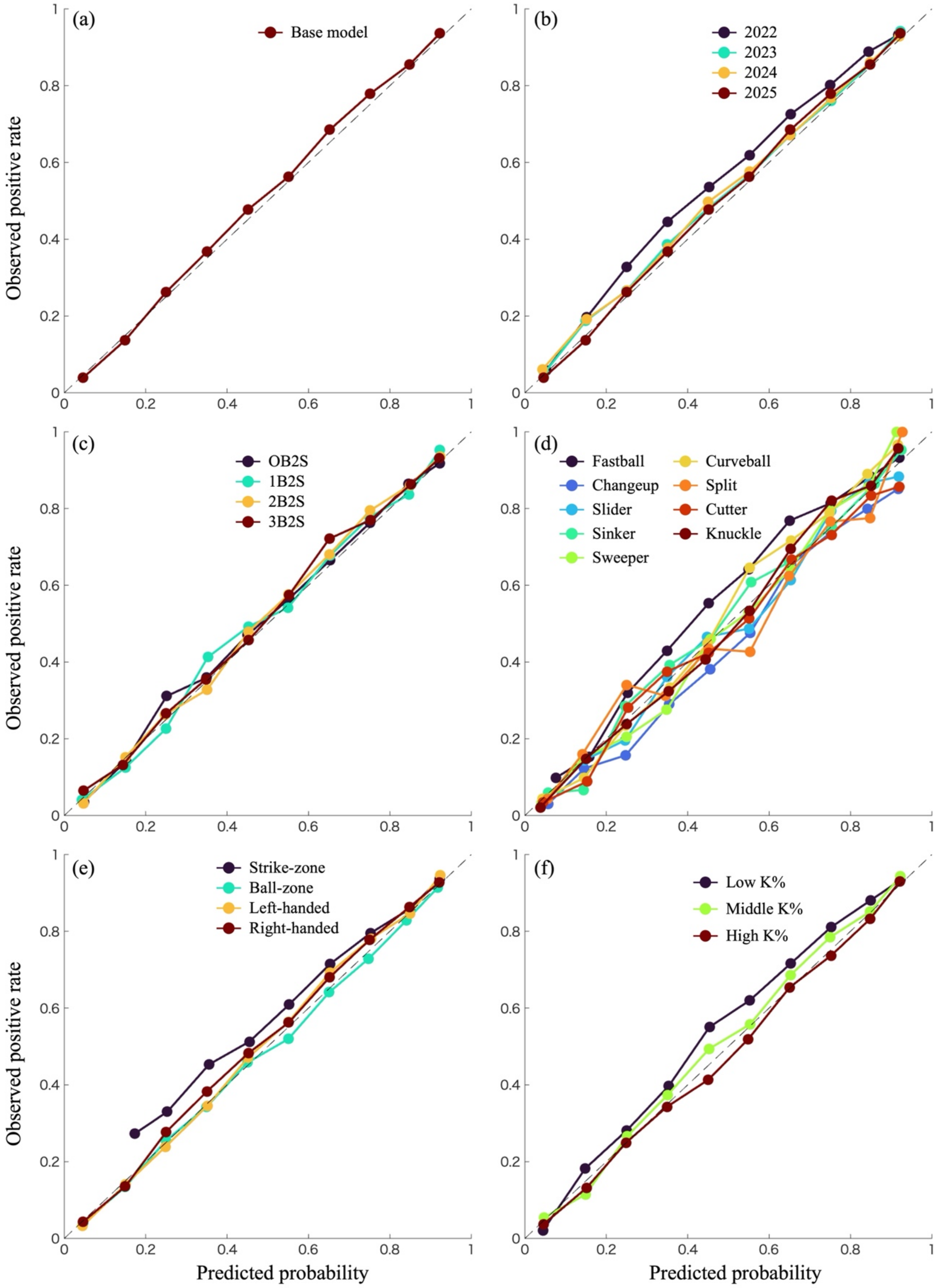

Figure 3. Calibration curves for each factor: (a) Base model, (b) season, (c) ball count, (d) pitch type, (e) strike-zone location and batter handedness, and (f) pitcher K%. Panels (a) and (c–f) show the mean values obtained when the 2025 season was used as the test dataset.

### 3.2 Analysis 1: Ablation study

Table 6 shows the results of the ablation study using the 2025 data as the test dataset. Results for the other seasons are provided in the Supplementary Information (Table S6–S8). Figure 4 shows the comparison of ROC AUC across training conditions.

As shown in the table and figure, the most critical factor affecting predictive performance was whether the final pitch was included. Although models incorporating the pitch sequence, such as the Full model and Last1–4 conditions, showed better predictive performance than the Last1 condition, the differences were less than 0.01. Considering the confidence intervals, these results do not provide clear evidence that including preceding pitch information improved predictive performance.

In a more detailed comparison across models, conditions using only Last2, Last3, Last4, Last5, or Last6 still showed better predictive performance than the constant baseline. However, their performance was comparable to that of the context-only condition, suggesting that this improvement was mainly attributable to the contextual information also included in the input. The similar predictive performance of the partial-shuffle condition and the Full model suggests that the order of the preceding pitches also had little influence on predictive performance. However, because the all-shuffle model outperformed the context-only condition, the overall distribution of pitches including the final pitch may contain information for distinguishing whether a plate appearance results in a ball put into play. Finally, compared with the simpler logistic regression baseline, the Transformer model showed better predictive performance in each condition.

These results were consistent across conditions using different seasons as the test datasets (Tables S6–S8).

Table 6. Results of the ablation study comparing predictive performance across training conditions using the 2025 data as the test dataset.

| Model Name | Accuracy | Precision | Recall | F1-score | ROC AUC |
|---|---|---|---|---|---|
| Full | 0.750 (0.743–0.757) | 0.763 (0.755–0.772) | 0.845 (0.838–0.852) | 0.802 (0.796–0.808) | 0.804 (0.797–0.811) |
| Last1 | 0.744 (0.737–0.750) | 0.761 (0.752–0.768) | 0.836 (0.829–0.844) | 0.796 (0.790–0.802) | 0.804 (0.797–0.811) |
| Last2 | 0.601 (0.593–0.610) | 0.609 (0.602–0.618) | 0.932 (0.927–0.937) | 0.737 (0.731–0.744) | 0.586 (0.577–0.595) |
| Last3 | 0.605 (0.597–0.613) | 0.611 (0.603–0.619) | 0.940 (0.935–0.944) | 0.740 (0.734–0.747) | 0.590 (0.580–0.599) |
| Last4 | 0.603 (0.596–0.611) | 0.611 (0.603–0.619) | 0.929 (0.924–0.935) | 0.737 (0.731–0.743) | 0.587 (0.578–0.597) |
| Last5 | 0.602 (0.594–0.610) | 0.607 (0.600–0.616) | 0.950 (0.946–0.954) | 0.741 (0.735–0.748) | 0.582 (0.573–0.591) |
| Last6 | 0.604 (0.596–0.612) | 0.607 (0.599–0.615) | 0.962 (0.959–0.966) | 0.744 (0.738–0.751) | 0.583 (0.575–0.594) |
| Last1–2 | 0.745 (0.739–0.752) | 0.762 (0.754–0.770) | 0.837 (0.830–0.845) | 0.798 (0.792–0.804) | 0.802 (0.794–0.809) |
| Last1–3 | 0.747 (0.740–0.753) | 0.761 (0.753–0.769) | 0.843 (0.835–0.850) | 0.800 (0.794–0.806) | 0.804 (0.797–0.811) |
| Last1–4 | 0.749 (0.742–0.755) | 0.764 (0.756–0.772) | 0.841 (0.833–0.848) | 0.801 (0.795–0.806) | 0.806 (0.799–0.813) |
| Last1–5 | 0.748 (0.742–0.754) | 0.763 (0.755–0.771) | 0.840 (0.833–0.847) | 0.800 (0.794–0.806) | 0.804 (0.797–0.811) |
| Without final pitch | 0.604 (0.596–0.611) | 0.614 (0.606–0.622) | 0.914 (0.909–0.919) | 0.734 (0.729–0.741) | 0.583 (0.574–0.592) |
| All-shuffle | 0.662 (0.654–0.669) | 0.676 (0.668–0.685) | 0.837 (0.829–0.844) | 0.748 (0.741–0.754) | 0.684 (0.675–0.692) |
| Partial-shuffle | 0.749 (0.743–0.756) | 0.763 (0.756–0.771) | 0.844 (0.837–0.851) | 0.801 (0.796–0.807) | 0.803 (0.796–0.810) |
| Final pitch plus one | 0.744 (0.737–0.751) | 0.760 (0.752–0.768) | 0.837 (0.830–0.844) | 0.796 (0.790–0.803) | 0.801 (0.794–0.808) |
| Context only | 0.601 (0.594–0.609) | 0.607 (0.599–0.615) | 0.949 (0.944–0.953) | 0.741 (0.734–0.747) | 0.583 (0.574–0.592) |
| Constant baseline | 0.600 (0.592–0.607) | 0.600 (0.592–0.607) | 1.000 (1.000–1.000) | 0.750 (0.744–0.756) | 0.500 (0.500–0.500) |
| Logistic regression | 0.660 (0.652–0.667) | 0.674 (0.666–0.683) | 0.837 (0.829–0.844) | 0.747 (0.740–0.753) | 0.640 (0.631–0.649) |

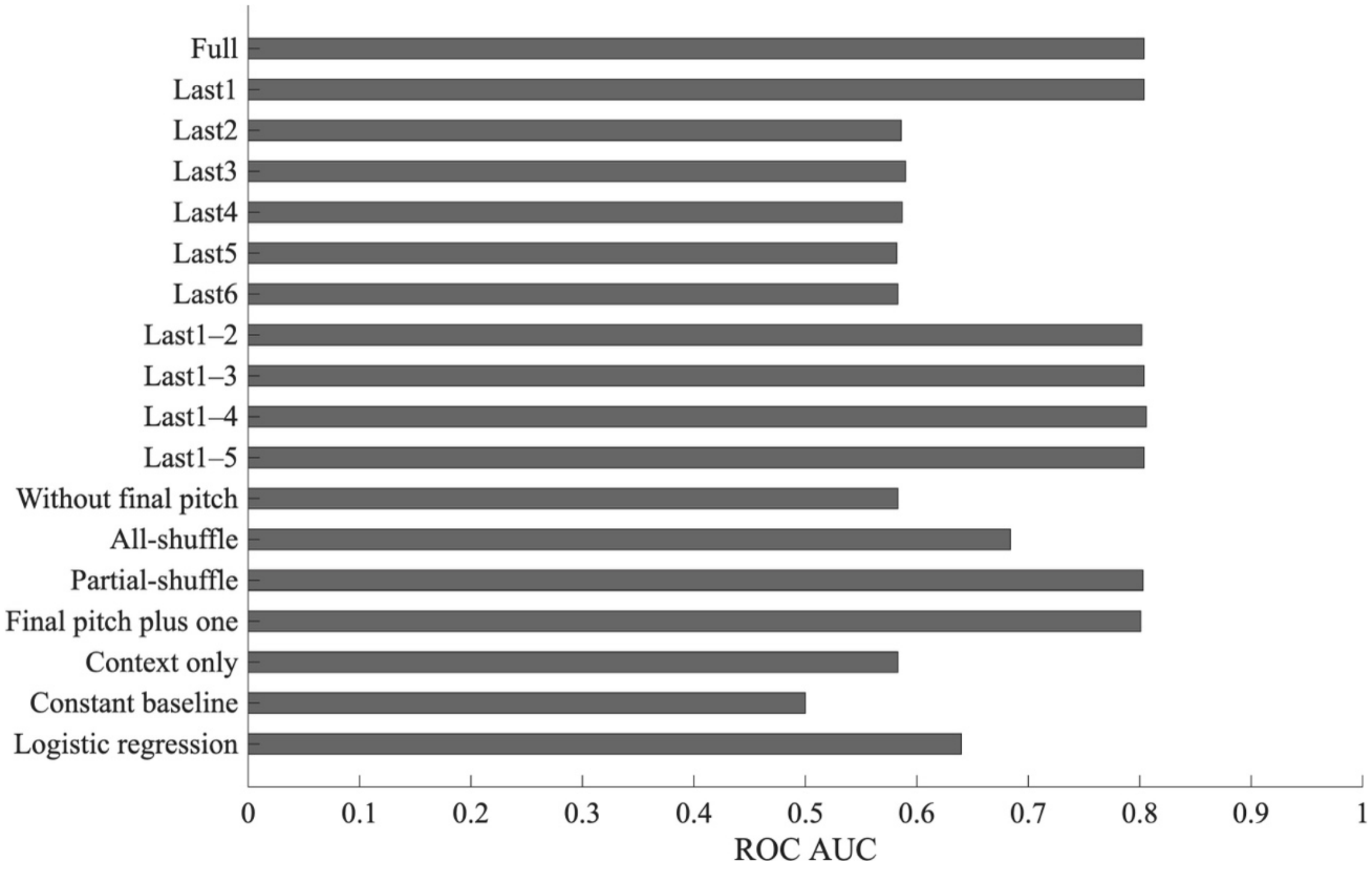

Figure 4. Comparison of ROC AUC across training conditions using the 2025 data as the test dataset.

### 3.3 Analysis 2: Replacement analysis

Figure 5 shows the results of the replacement analysis. The mean magnitudes of change in model output across the three seasons were 0.451, 0.144, 0.137, 0.139, 0.128, and 0.125 for replacement of the last 1–6 pitches for observed-pitch condition (Figure 5 (a)); 0.449, 0.134, 0.129, 0.130, 0.120, and 0.117 for artificial-pitch condition (Figure 5 (b)); 0.173, 0.051, 0.050, 0.051, 0.046 and 0.046 for observed-pitch condition with pitch-type replacement only (Figure 5 (c)); 0.210, 0.061, 0.061, 0.061, 0.056 and 0.054 for artificial-pitch condition with pitch-type replacement only (Figure 5 (d)); 0.240, 0.068, 0.064, 0.065, 0.060 and 0.058 for observed-pitch condition with location replacement only (Figure 5 (e)), and 0.243, 0.063, 0.058, 0.059, 0.055 and 0.053 for artificial-pitch condition with location replacement only (Figure 5 (f)).

Overall, the effect on model output of replacing preceding pitches within the ball zone was smaller than that of replacing the final pitch within the strike zone, with the magnitude of change being approximately 26–32%. This trend was consistently observed across seasons and candidate-pitch generation conditions (observed or artificial). Among the preceding pitches, pitches closer to the final pitch tended to have a slightly greater influence on model output. In the attribute analysis, within the range of replacements examined in this study, changes in pitch location produced larger changes in model output than changes in pitch type.

Because the change in model output caused by replacing preceding pitches was approximately 0.11–0.14, it rarely changed whether the predicted probability crossed the binary classification threshold of 0.50. This may explain why preceding pitches had only a small effect on predictive performance in Analysis 1.

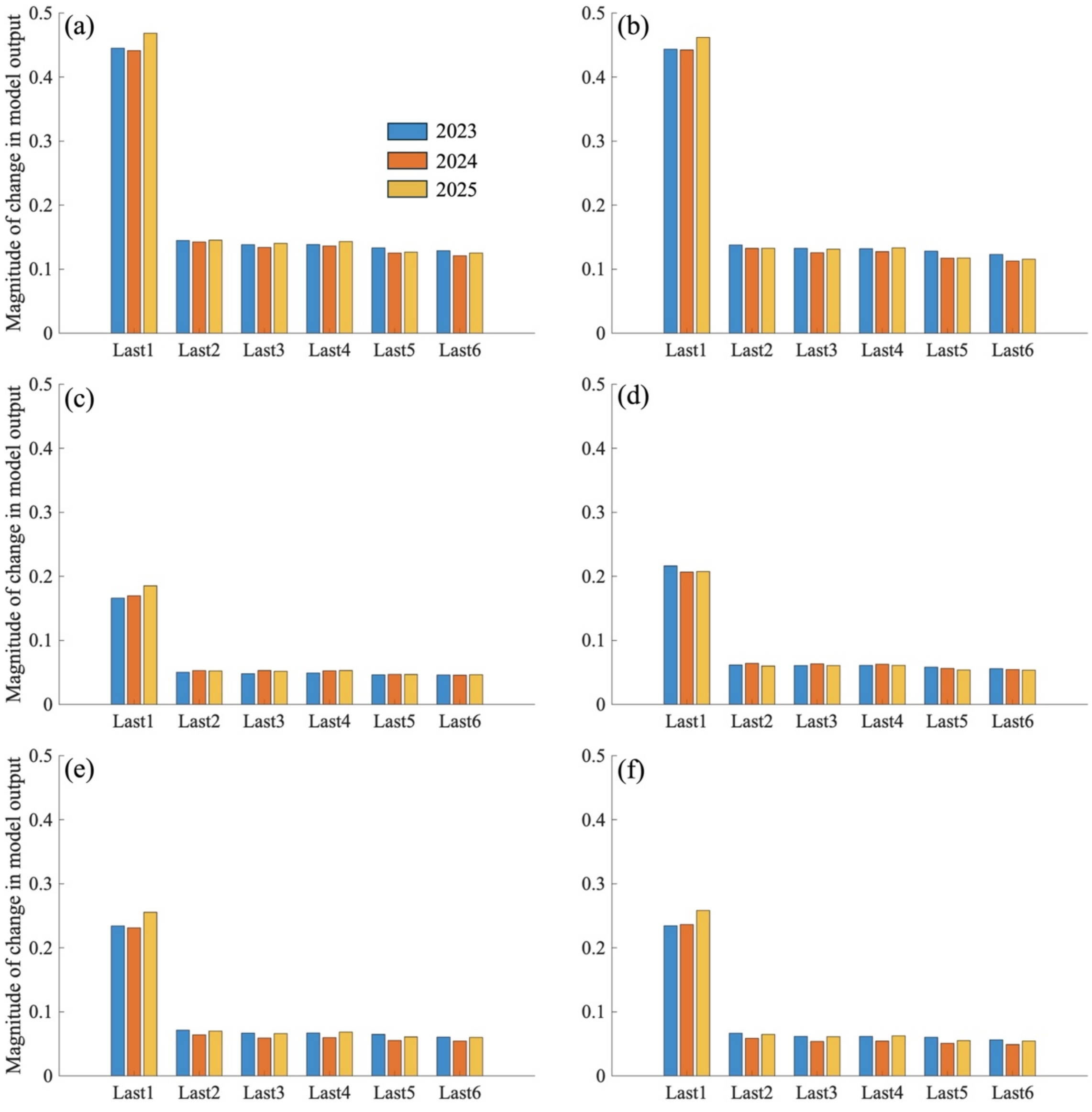

Figure 5. Results of the replacement analysis: (a) observed-pitch condition, (b) artificial-pitch condition, (c) observed-pitch condition with pitch-type replacement only, (d) artificial-pitch condition with pitch-type replacement only, (e) observed-pitch condition with location replacement only, and (f) artificial-pitch condition with location replacement only.

## 4. Discussion

The results of this study suggest that the final pitch makes the greatest contribution to the model's prediction of hitting outcomes, whereas the contribution of preceding pitches is relatively small. Therefore, preceding pitches played only a supplementary role in slightly modifying the predicted in-play probability of the ML model. This finding seemingly contradicts practical knowledge emphasizing the importance of preceding pitches.

There are several potential explanations for these results. First, information from the preceding pitch sequence may have been compressed into the final pitch. Given that the selection of a pitch in a given situation considers the pitch history, the preceding sequence may be inferred from the final pitch. In this case, even if preceding pitches are practically important, predictions may still be possible using only the final pitch. The improvement in predictive performance observed in the all-shuffle model (Figure 4) may also suggest that pitch order was partly decoded from the entire pitch distribution. Therefore, based on the present findings, future studies should examine these effects under more controlled conditions, for example using batting tasks in virtual reality environments (e.g., Hayashi et al., 2026) with randomly generated pitch sequences.

Moreover, the binary prediction task (i.e., in-play vs swinging strikeout) used in this study may also have influenced the results. A swinging miss generally results from a relatively large spatiotemporal error, whereas the quality of contact, such as exit velocity, can vary with much smaller errors (Kidokoro et al., 2019; Kidokoro and Morishita 2021). Given the fact that previous experience induces perceptual and motor biases in hitting movements (Gray and Cañal-Bruland, 2018; Takamido et al., 2021; Nakamoto et al., 2022), different results might be obtained if the quality of batted-ball outcomes (e.g., expected slugging value) were used as the prediction target.

Finally, the relatively short sequence of up to six pitches may have limited the Transformer's ability to learn temporal dependencies. Batters may consider not only the sequence within the current plate appearance but also previous plate appearances and past matchups. Therefore, as in the previous study that considered longer sequences of up to 400 pitches (Kneita, 2025), incorporating longer pitch histories may reduce the model's dependence on the final pitch.

Therefore, considering these factors, the present findings should be interpreted as an initial step toward quantifying the importance of preceding pitches, and conclusions should be drawn after these considerations have been further investigated.

From a practical perspective, these results should be interpreted with caution. Because even a 10% difference in hitting success can be practically meaningful (e.g., a batting average of 0.200 vs. 0.300), the small influence of preceding pitches on the model's predicted probability may still have practical significance if it corresponds to actual differences in hitting outcomes. Indeed, considering the calibration results of the model (Figure 3 and Table 5), the changes in output caused by replacing preceding pitches may not be negligible. Therefore, to accurately understand their role and importance, hypothesis-driven experiments should be conducted under controlled conditions such as those described above.

This study had several limitations. First, because this study focused only on hit-into-play outcomes (or swinging strikes), future studies should develop a more comprehensive model incorporating balls, called strikes, fouls, batted-ball outcomes, subsequent pitch selection, and hitter adaptation across a wider range of game situations other than two-strike counts. In addition, because the replacement analysis was restricted to final pitches within the strike zone, where batters were assumed to swing, future studies should incorporate the probability that the batter attempts to swing, which would allow pitches outside the strike zone, such as chase pitches, to be examined. Moreover, a pitch observed for a given pitcher was assumed to remain executable across different counts, handedness matchups, and game situations, and realized rather than intended pitch location was used without accounting for pitcher-specific command; these issues should also be addressed in future studies. Finally, because this study was limited to examining the contribution of preceding pitches to prediction, future studies should also investigate which specific pitch combinations are most effective to deepen the understanding of baseball strategy.

## 5. Conclusion

In conclusion, the analyses in this study suggest that preceding pitches contributed less to in-play prediction than the final pitch, and their inclusion did not improve the predictive performance of the ML model beyond the range of the confidence intervals. However, preceding pitches had a small but potentially non-negligible effect on the model's probability output when the calibration results were considered. Because this small effect may be practically meaningful, based on the results of this study, future studies are expected to conduct more detailed and rigorous investigations to better understand the role of preceding pitches in baseball strategy. The contribution of this study is to first quantify the importance of preceding pitches for hitting outcome prediction and provide the basis for future studies.

**Supplementary Information**

**1. Supplementary tables: Sample selection, descriptive summaries by attribute, and results not included in the main text.**

Table S1. Sample selection and exclusion process for each year. The numbers indicate the number of pitches excluded at each step (not plate appearances).

| Year | Process name | Input sample | Output sample | Exclude sample |
|---|---|---|---|---|
| 2018 | Target pitcher selection | 734567 | 181406 | 553161 |
| | Two-strike outcome selection | 181406 | 87868 | 93538 |
| | Ball 1 pitch before final pitch | 87868 | 31494 | 56374 |
| | Ball 2 pitches before final pitch | 87868 | 31144 | 56724 |
| | Ball 3 pitches before final pitch | 87868 | 32208 | 55660 |
| | Ball 4 pitches before final pitch | 87868 | 26385 | 61483 |
| | Ball 5 pitches before final pitch | 87868 | 17191 | 70677 |
| 2019 | Target pitcher selection | 763198 | 196396 | 566802 |
| | Two-strike outcome selection | 196396 | 93962 | 102434 |
| | Ball 1 pitch before final pitch | 93962 | 34253 | 59709 |
| | Ball 2 pitches before final pitch | 93962 | 33559 | 60403 |
| | Ball 3 pitches before final pitch | 93962 | 33990 | 59972 |
| | Ball 4 pitches before final pitch | 93962 | 28410 | 65552 |
| | Ball 5 pitches before final pitch | 93962 | 17954 | 76008 |
| 2022 | Target pitcher selection | 775330 | 137961 | 637369 |
| | Two-strike outcome selection | 137961 | 72782 | 65179 |
| | Ball 1 pitch before final pitch | 72782 | 25805 | 46977 |
| | Ball 2 pitches before final pitch | 72782 | 24843 | 47939 |
| | Ball 3 pitches before final pitch | 72782 | 26187 | 46595 |
| | Ball 4 pitches before final pitch | 72782 | 21570 | 51212 |
| | Ball 5 pitches before final pitch | 72782 | 13704 | 59078 |
| 2023 | Target pitcher selection | 774038 | 142586 | 631452 |
| | Two-strike outcome selection | 142586 | 71294 | 71292 |
| | Ball 1 pitch before final pitch | 71294 | 26966 | 71292 |
| | Ball 2 pitches before final pitch | 71294 | 25665 | 44328 |
| | Ball 3 pitches before final pitch | 71294 | 25269 | 45629 |
| | Ball 4 pitches before final pitch | 71294 | 21628 | 46025 |
| | Ball 5 pitches before final pitch | 71294 | 13800 | 57494 |
| 2024 | Target pitcher selection | 760249 | 179457 | 580792 |
| | Two-strike outcome selection | 179457 | 90556 | 88901 |
| | Ball 1 pitch before final pitch | 90556 | 33717 | 56839 |
| | Ball 2 pitches before final pitch | 90556 | 32857 | 57699 |
| | Ball 3 pitches before final pitch | 90556 | 32166 | 58390 |
| | Ball 4 pitches before final pitch | 90556 | 27315 | 63241 |
| | Ball 5 pitches before final pitch | 90556 | 17722 | 72834 |
| 2025 | Target pitcher selection | 770795 | 164493 | 606302 |
| | Two-strike outcome selection | 164493 | 82083 | 82410 |
| | Ball 1 pitch before final pitch | 82083 | 30442 | 51641 |
| | Ball 2 pitches before final pitch | 82083 | 28976 | 53107 |
| | Ball 3 pitches before final pitch | 82083 | 29686 | 52397 |
| | Ball 4 pitches before final pitch | 82083 | 25326 | 56757 |
| | Ball 5 pitches before final pitch | 82083 | 16442 | 65641 |

Table S2. Number of target plate appearances, pitchers, and plate appearances by batter handedness included in the dataset for each season.

| Year | Plate appearances | Pitchers | Right-handed batters | Light-handed batters |
|---|---|---|---|---|
| 2018 | 17756 | 57 | 10578 | 7178 |
| 2019 | 18996 | 61 | 11016 | 7980 |
| 2022 | 14822 | 45 | 8911 | 5911 |
| 2023 | 14399 | 44 | 8257 | 6142 |
| 2024 | 18203 | 58 | 10764 | 7439 |
| 2025 | 16487 | 52 | 9593 | 6894 |

Table S3. Summary of the number of pitches for each pitch type (CH: Changeup, CS: Slow Curve, CU: Curveball, EP: Eephus, FA: Other, FC: Cutter, FO: Forkball, FS: Split-Finger, KC: Knuckle Curve, KN: Knuckleball, PO: Pitch Out, SI: Sinker, SL: Slider, ST: Sweeper, SV: Slurve, UN: Unknown). FA, UN, and PO were excluded from subsequent analyses.

| Year | CH | CS | CU | EP | FA | FC | FF | FO | FS |
|---|---|---|---|---|---|---|---|---|---|
| 2018 | 10765 | 0 | 8229 | 141 | 0 | 5221 | 30048 | 0 | 1008 |
| 2019 | 10220 | 4 | 9333 | 20 | 0 | 7243 | 32600 | 0 | 1470 |
| 2022 | 8799 | 28 | 7668 | 0 | 0 | 6911 | 22254 | 0 | 1196 |
| 2023 | 7840 | 3 | 6626 | 0 | 1 | 5162 | 23225 | 410 | 1679 |
| 2024 | 11697 | 16 | 6654 | 0 | 0 | 7125 | 27239 | 0 | 3059 |
| 2025 | 10933 | 59 | 5857 | 0 | 0 | 5063 | 26094 | 0 | 3083 |
| Year | KC | KN | PO | SI | SL | ST | SV | UN | |
| 2018 | 3642 | 0 | 4 | 14446 | 12496 | 1313 | 546 | 9 | |
| 2019 | 3559 | 0 | 9 | 13475 | 13459 | 1389 | 1181 | 0 | |
| 2022 | 3856 | 0 | 3 | 9584 | 8738 | 2414 | 1331 | 0 | |
| 2023 | 2161 | 2 | 2 | 9424 | 9711 | 4536 | 512 | 0 | |
| 2024 | 2464 | 0 | 4 | 13187 | 10103 | 8270 | 737 | 1 | |
| 2025 | 2638 | 0 | 6 | 12082 | 9232 | 6662 | 374 | 0 | |

Table S4. Number of target plate appearances for each count by year.

| Year | 0B-2S | 1B-2S | 2B-2S | 3B-2S |
|---|---|---|---|---|
| 2018 | 3511 | 5690 | 5165 | 3390 |
| 2019 | 3858 | 5924 | 5569 | 3645 |
| 2022 | 3066 | 4844 | 4350 | 2562 |
| 2023 | 2848 | 4520 | 4252 | 2779 |
| 2024 | 3429 | 5865 | 5419 | 3490 |
| 2025 | 3193 | 5170 | 4917 | 3207 |

Table S5. Distribution of outcomes for target plate appearances by year.

| Year | Strikeout | Out | Single | Double |
|---|---|---|---|---|
| 2018 | 7016 | 7286 | 2231 | 649 |
| 2019 | 7517 | 7670 | 2448 | 668 |
| 2022 | 5800 | 6261 | 1827 | 499 |
| 2023 | 5956 | 5637 | 1809 | 520 |
| 2024 | 7198 | 7452 | 2338 | 629 |
| 2025 | 6601 | 6733 | 2053 | 562 |

| Year | Triple | Homerun | Error | Fielders choice |
|---|---|---|---|---|
| 2018 | 57 | 379 | 115 | 23 |
| 2019 | 77 | 446 | 141 | 29 |
| 2022 | 36 | 301 | 77 | 21 |
| 2023 | 55 | 320 | 88 | 14 |
| 2024 | 66 | 388 | 104 | 28 |
| 2025 | 46 | 381 | 86 | 25 |

Table S6. Results of the ablation study comparing predictive performance across training conditions using the 2022 data as the test dataset.

| Model Name | Accuracy | Precision | Recall | F1-score | ROC AUC |
|---|---|---|---|---|---|
| Full | 0.733 (0.726–0.740) | 0.774 (0.766–0.783) | 0.793 (0.784–0.800) | 0.783 (0.777–0.790) | 0.791 (0.784–0.799) |
| Last1 | 0.735 (0.727–0.742) | 0.780 (0.772–0.789) | 0.785 (0.777–0.793) | 0.783 (0.776–0.790) | 0.795 (0.787–0.802) |
| Last2 | 0.590 (0.582–0.598) | 0.634 (0.625–0.643) | 0.774 (0.765–0.782) | 0.697 (0.689–0.704) | 0.569 (0.560–0.579) |
| Last3 | 0.585 (0.578–0.593) | 0.631 (0.622–0.640) | 0.768 (0.759–0.777) | 0.693 (0.686–0.700) | 0.564 (0.554–0.574) |
| Last4 | 0.593 (0.585–0.600) | 0.626 (0.618–0.634) | 0.824 (0.816–0.832) | 0.711 (0.705–0.718) | 0.563 (0.553–0.571) |
| Last5 | 0.595 (0.587–0.602) | 0.622 (0.614–0.630) | 0.853 (0.846–0.860) | 0.719 (0.712–0.726) | 0.558 (0.549–0.567) |
| Last6 | 0.595 (0.587–0.603) | 0.619 (0.611–0.628) | 0.869 (0.862–0.876) | 0.723 (0.716–0.730) | 0.562 (0.553–0.571) |
| Last1–2 | 0.733 (0.726–0.741) | 0.776 (0.767–0.785) | 0.789 (0.781–0.798) | 0.783 (0.776–0.789) | 0.791 (0.784–0.799) |
| Last1–3 | 0.731 (0.724–0.739) | 0.774 (0.766–0.783) | 0.788 (0.780–0.796) | 0.781 (0.775–0.788) | 0.792 (0.784–0.800) |
| Last1–4 | 0.731 (0.724–0.738) | 0.776 (0.767–0.784) | 0.784 (0.776–0.793) | 0.780 (0.773–0.787) | 0.786 (0.779–0.794) |
| Last1–5 | 0.732 (0.725–0.739) | 0.775 (0.766–0.784) | 0.788 (0.779–0.797) | 0.782 (0.775–0.788) | 0.790 (0.782–0.797) |
| Without final pitch | 0.591 (0.583–0.599) | 0.635 (0.626–0.643) | 0.775 (0.766–0.783) | 0.698 (0.690–0.705) | 0.566 (0.557–0.576) |
| All-shuffle | 0.637 (0.629–0.645) | 0.686 (0.677–0.695) | 0.744 (0.734–0.754) | 0.714 (0.706–0.721) | 0.659 (0.649–0.668) |
| Partial-shuffle | 0.736 (0.729–0.743) | 0.777 (0.769–0.785) | 0.795 (0.786–0.803) | 0.786 (0.779–0.792) | 0.794 (0.787–0.802) |
| Final pitch plus one | 0.735 (0.727–0.741) | 0.774 (0.765–0.782) | 0.797 (0.789–0.805) | 0.785 (0.779–0.792) | 0.788 (0.781–0.796) |
| Context only | 0.583 (0.575–0.590) | 0.635 (0.626–0.644) | 0.739 (0.729–0.748) | 0.683 (0.676–0.691) | 0.572 (0.563–0.582) |
| Constant baseline | 0.609 (0.601–0.617) | 0.609 (0.601–0.617) | 1.000 (1.000–1.000) | 0.757 (0.751–0.763) | 0.500 (0.500–0.500) |
| Logistic regression | 0.633 (0.626–0.640) | 0.688 (0.678–0.697) | 0.727 (0.718–0.736) | 0.707 (0.700–0.714) | 0.637 (0.628–0.646) |

Table S7. Results of the ablation study comparing predictive performance across training conditions

using the 2023 data as the test dataset.

| Model Name | Accuracy | Precision | Recall | F1-score | ROC AUC |
|---|---|---|---|---|---|
| Full | 0.751 (0.744–0.758) | 0.762 (0.753–0.771) | 0.837 (0.829–0.845) | 0.798 (0.791–0.804) | 0.814 (0.806–0.821) |
| Last1 | 0.751 (0.744–0.758) | 0.755 (0.747–0.764) | 0.850 (0.843–0.858) | 0.800 (0.794–0.806) | 0.814 (0.807–0.821) |
| Last2 | 0.592 (0.584–0.601) | 0.603 (0.594–0.611) | 0.894 (0.888–0.901) | 0.720 (0.713–0.727) | 0.588 (0.579–0.597) |
| Last3 | 0.593 (0.585–0.601) | 0.603 (0.595–0.613) | 0.891 (0.884–0.897) | 0.719 (0.713–0.727) | 0.582 (0.573–0.591) |
| Last4 | 0.587 (0.579–0.596) | 0.597 (0.589–0.606) | 0.909 (0.902–0.915) | 0.721 (0.714–0.728) | 0.581 (0.572–0.590) |
| Last5 | 0.588 (0.580–0.596) | 0.592 (0.584–0.600) | 0.957 (0.952–0.961) | 0.731 (0.725–0.738) | 0.576 (0.567–0.585) |
| Last6 | 0.589 (0.580–0.597) | 0.593 (0.585–0.601) | 0.953 (0.948–0.957) | 0.731 (0.724–0.738) | 0.579 (0.570–0.588) |
| Last1–2 | 0.745 (0.738–0.752) | 0.752 (0.744–0.761) | 0.843 (0.835–0.851) | 0.795 (0.789–0.801) | 0.809 (0.801–0.815) |
| Last1–3 | 0.750 (0.743–0.757) | 0.757 (0.749–0.766) | 0.845 (0.837–0.853) | 0.799 (0.792–0.805) | 0.813 (0.805–0.819) |
| Last1–4 | 0.751 (0.744–0.758) | 0.764 (0.756–0.773) | 0.832 (0.824–0.840) | 0.797 (0.790–0.803) | 0.813 (0.805–0.819) |
| Last1–5 | 0.751 (0.744–0.758) | 0.762 (0.754–0.771) | 0.837 (0.828–0.845) | 0.798 (0.791–0.804) | 0.813 (0.806–0.820) |
| Without final pitch | 0.590 (0.582–0.598) | 0.607 (0.598–0.615) | 0.856 (0.848–0.863) | 0.710 (0.703–0.717) | 0.582 (0.572–0.591) |
| All-shuffle | 0.661 (0.653–0.668) | 0.675 (0.666–0.684) | 0.815 (0.807–0.823) | 0.738 (0.731–0.745) | 0.689 (0.680–0.697) |
| Partial-shuffle | 0.748 (0.741–0.756) | 0.760 (0.751–0.768) | 0.834 (0.826–0.842) | 0.795 (0.788–0.802) | 0.813 (0.805–0.819) |
| Final pitch plus one | 0.745 (0.738–0.752) | 0.756 (0.748–0.765) | 0.834 (0.825–0.841) | 0.793 (0.786–0.799) | 0.807 (0.800–0.815) |
| Context only | 0.592 (0.584–0.600) | 0.601 (0.593–0.610) | 0.902 (0.895–0.908) | 0.721 (0.715–0.728) | 0.587 (0.578–0.597) |
| Constant baseline | 0.586 (0.579–0.595) | 0.586 (0.579–0.595) | 1.000 (1.000–1.000) | 0.739 (0.733–0.746) | 0.500 (0.500–0.500) |
| Logistic regression | 0.661 (0.654–0.669) | 0.668 (0.659–0.677) | 0.839 (0.831–0.847) | 0.744 (0.737–0.751) | 0.649 (0.639–0.658) |

Table S8. Results of the ablation study comparing predictive performance across training conditions using the 2024 data as the test dataset.

| Model Name | Accuracy | Precision | Recall | F1-score | ROC AUC |
|---|---|---|---|---|---|
| Full | 0.744 (0.738–0.751) | 0.758 (0.750–0.765) | 0.848 (0.842–0.855) | 0.800 (0.795–0.806) | 0.795 (0.789–0.802) |
| Last1 | 0.742 (0.736–0.749) | 0.757 (0.749–0.764) | 0.846 (0.839–0.852) | 0.799 (0.793–0.805) | 0.794 (0.787–0.800) |
| Last2 | 0.604 (0.597–0.611) | 0.621 (0.613–0.629) | 0.886 (0.880–0.892) | 0.730 (0.724–0.736) | 0.575 (0.567–0.584) |
| Last3 | 0.605 (0.597–0.612) | 0.621 (0.613–0.628) | 0.889 (0.883–0.894) | 0.731 (0.725–0.737) | 0.577 (0.568–0.586) |
| Last4 | 0.604 (0.597–0.611) | 0.620 (0.612–0.628) | 0.889 (0.883–0.895) | 0.731 (0.725–0.736) | 0.578 (0.569–0.587) |
| Last5 | 0.604 (0.597–0.611) | 0.618 (0.611–0.626) | 0.902 (0.895–0.907) | 0.734 (0.727–0.739) | 0.574 (0.566–0.582) |
| Last6 | 0.609 (0.602–0.615) | 0.616 (0.608–0.623) | 0.939 (0.935–0.943) | 0.744 (0.738–0.749) | 0.578 (0.569–0.586) |
| Last1–2 | 0.741 (0.735–0.747) | 0.757 (0.750–0.765) | 0.841 (0.835–0.848) | 0.797 (0.791–0.803) | 0.793 (0.786–0.800) |
| Last1–3 | 0.742 (0.735–0.748) | 0.755 (0.748–0.763) | 0.848 (0.841–0.854) | 0.799 (0.793–0.805) | 0.796 (0.789–0.803) |
| Last1–4 | 0.744 (0.738–0.750) | 0.759 (0.751–0.766) | 0.845 (0.838–0.852) | 0.799 (0.794–0.805) | 0.795 (0.789–0.802) |
| Last1–5 | 0.743 (0.736–0.749) | 0.757 (0.749–0.764) | 0.846 (0.840–0.853) | 0.799 (0.793–0.805) | 0.795 (0.789–0.802) |
| Without final pitch | 0.603 (0.596–0.610) | 0.621 (0.613–0.629) | 0.881 (0.875–0.887) | 0.728 (0.722–0.735) | 0.579 (0.571–0.588) |
| All-shuffle | 0.656 (0.649–0.663) | 0.681 (0.672–0.689) | 0.812 (0.805–0.819) | 0.741 (0.734–0.747) | 0.677 (0.669–0.685) |
| Partial-shuffle | 0.740 (0.734–0.747) | 0.753 (0.745–0.761) | 0.848 (0.841–0.855) | 0.798 (0.792–0.804) | 0.793 (0.787–0.800) |
| Final pitch plus one | 0.742 (0.735–0.748) | 0.754 (0.747–0.762) | 0.849 (0.843–0.856) | 0.799 (0.793–0.805) | 0.792 (0.785–0.799) |
| Context only | 0.606 (0.599–0.612) | 0.622 (0.614–0.629) | 0.888 (0.882–0.894) | 0.731 (0.726–0.737) | 0.578 (0.570–0.587) |
| Constant baseline | 0.605 (0.597–0.612) | 0.605 (0.597–0.612) | 1.000 (1.000–1.000) | 0.754 (0.748–0.759) | 0.500 (0.500–0.500) |
| Logistic regression | 0.653 (0.646–0.660) | 0.675 (0.667–0.683) | 0.820 (0.813–0.827) | 0.741 (0.735–0.746) | 0.625 (0.616–0.633) |

## 2. Hyperparameter tuning of the Transformer model

This Supplementary Information file describes the hyperparameter-tuning procedure used for the Transformer architecture. Because the Transformer includes several hyperparameters owing to its architectural complexity, hyperparameter tuning was performed using the following procedure in this study.

The search space included the Transformer embedding dimension, number of attention heads, feed-forward dimension, number of Transformer Encoder blocks, and the number of units in the dense layer after concatenation. Table S9 shows the candidate values for each hyperparameter. As shown in the table, the hyperparameter candidates were designed to cover a broad range of model capacities, from lightweight to larger-scale models. The dense layer size of the contextual-information processing pathway, learning rate, and batch size were fixed at 32, $1\times10^{-4}$, and 512, respectively. Each candidate model was trained for up to 200 epochs with early stopping based on validation AUC, using a patience of 10 epochs. The best model was selected by maximizing validation AUC, with validation loss used as a secondary criterion.

Table S10 shows the results of the hyperparameter tuning. Based on the validation AUC of each candidate model, Model 6 was adopted as the base model used in the analysis of this study.

Table S9. Hyperparameter candidates for the Transformer model

| Model | Embedding dimension | Number of heads | Feedforward dimension | Encoder blocks | Dense units |
|---|---|---|---|---|---|
| 1 | 64 | 2 | 128 | 1 | 64 |
| 2 | 64 | 4 | 128 | 1 | 64 |
| 3 | 64 | 4 | 256 | 2 | 64 |
| 4 | 128 | 4 | 256 | 2 | 64 |
| 5 | 128 | 4 | 512 | 2 | 64 |
| 6 | 256 | 4 | 512 | 2 | 64 |
| 7 | 256 | 8 | 512 | 2 | 64 |
| 8 | 256 | 8 | 1024 | 2 | 128 |

Table S10. The results of the hyperparameter tuning

| Model | Validation AUC | Model | Validation AUC |
|---|---|---|---|
| 1 | 0.802 | 5 | 0.805 |
| 2 | 0.801 | **6** | **0.808** |
| 3 | 0.807 | 7 | 0.806 |
| 4 | 0.804 | 8 | 0.807 |